\pdfoutput=1

\documentclass[11pt]{article}
\usepackage[]{emnlp2021}
\usepackage{times}
\usepackage{latexsym}
\usepackage[T1]{fontenc}
\usepackage[utf8]{inputenc}
\usepackage{microtype}
\usepackage{booktabs}

\usepackage[noend]{algpseudocode}
\usepackage{algorithmicx,algorithm}
\usepackage{multirow}
\usepackage{amsmath}

\usepackage{microtype}
\usepackage{hyperref}
\usepackage{graphics}
\usepackage{graphicx}

\newcommand*\samethanks[1][\value{footnote}]{\footnotemark[#1]}

\author{Adam Nik\textsuperscript{2\space 4}\thanks{\quad The two authors contributed equally to this work.}\space, Ge Zhang\textsuperscript{1\space 2\space 3}\samethanks\space, Xingran Chen\textsuperscript{3}, Mingyu Li\textsuperscript{2\space 3},  Jie Fu\thanks{\quad Corresponding Author}\; \textsuperscript{1}\\ \textsuperscript{1} Beijing Academy of Artificial Intelligence, China\\ 
\textsuperscript{2} 1Cademy Community, USA\\  
\textsuperscript{3} University of Michigan Ann Arbor, USA\\
\textsuperscript{4} Carleton College, USA\\
\texttt{\href{fujie@baai.ac.cn}{\color{black}{fujie AT baai.ac.cn}}}
}

\title{1Cademy @ Causal News Corpus 2022: Leveraging Self-Training in Causality Classification of Socio-Political Event Data}

\begin{document}
\maketitle
\begin{abstract}

This paper details our participation in the Challenges and Applications of Automated Extraction of Socio-political Events from Text (CASE) workshop @ EMNLP 2022, where we take part in Subtask 1 of Shared Task 3 \cite{tan-etal-2022-event}. 
We approach the given task of event causality detection by proposing a self-training pipeline that follows a teacher-student classifier method. 
More specifically, we initially train a teacher model on the true, original task data, and use that teacher model to self-label data to be used in the training of a separate student model for the final task prediction. 
We test how restricting the number of positive or negative self-labeled examples in the self-training process affects classification performance. 
Our final results show that using self-training produces a comprehensive performance improvement across all models and self-labeled training sets tested within the task of event causality sequence classification. On top of that, we find that self-training performance did not diminish even when restricting either positive/negative examples used in training.
Our code is be publicly available at \hyperlink{https://github.com/Gzhang-umich/1CademyTeamOfCASE}{https://github.com/Gzhang-umich/1CademyTeamOfCASE}.
\end{abstract}

\begin{figure*}
    \label{sec:figure1}
    \centering
    \includegraphics[width=\textwidth]{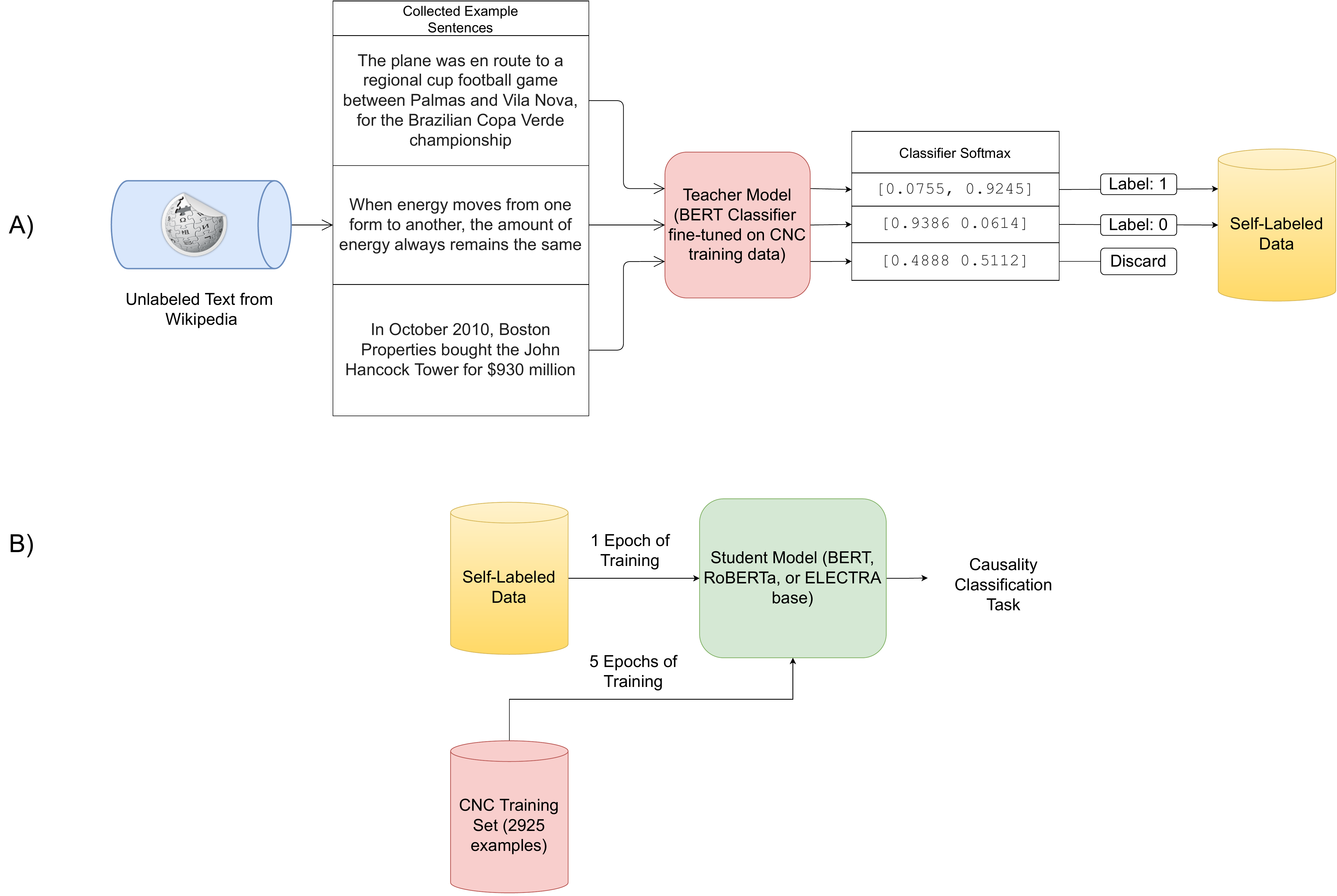}
    \captionsetup{justification=centering,margin=1cm}
    \caption{A) Self-training pipeline with Teacher Model. B) We use the self-labeled examples as part of the training when training in Student Models for the task of causality classification}
\end{figure*}

\section{Introduction}

Task 1 of the CASE workshop @ EMNLP 2022 works to identify and classify event causality in socio-political event (SPE) data, with subtask 1 being a binary classification of causality. In other words, participants are tasked with answering: Does an event sentence contain cause-effect meaning? The workshop provides data from Causal News Corpus (CNC) \cite{tan-EtAl:2022:LREC} for training and evaluation of the subtask. Causality itself aims to identify a semantic relationship between two events where one event (the cause) is responsible for the production of the other event (the effect). Utilizing the CNC dataset serves as a benchmark for participants to evaluate the ability of a given model or process to identify causality in event data.

We approach the problem of causality sequence classification by applying self-training \cite{ouali2020overview,van2020survey,10.1007/s10115-013-0706-y} as a means to improve the performance of language models in this task. 
The goal of self-training is to generate proxy labels for previously unlabeled data to enhance the learning process. The self-training process works by iteratively labeling previously unpredicted data, and then using the new pseudo-labels as truthful labels in the next training stage. The intuition behind self-training comes from the fact that it can pseudo-expand the training space to basically an unlimited size in a very cheap manner, as no hand-labeling is required in the process.

Additionally, we run supplementary experiments to test the effectiveness of self-training against various transformer-based data augmentation techniques \cite{feng2021survey} and separate multi-task learning approaches \cite{caruana1997multitask} that we originally designed for the competition. The description and results of these additional experiments can be found in the \hyperref[sec:appendix]{Appendix}.

In summary, our main contributions are as follows.

\textbf{1)} We propose a self-training pipeline for the task of causality detection in SPE data for the purposes of competing in Subtask 1 of Shared Task 3 of the CASE workshop @ EMNLP 2022. Our best model achieved 0.8135 accuracy and a 0.8398 F$_1$ score on the competition's test set.

\textbf{2)} We evaluate our self-training pipeline with collected self-labeled datasets of highly positive samples, highly negative samples, and even distributed positive and negative samples. We show that using self-labeled datasets improves performance across the board on all tested models, and that the performance increase provided by self-training did not significantly change based on the ratio of positive to negative self-labeled samples used in training.

For all implementations of our code, we use the HuggingFace Transformers library \cite{wolf-etal-2020-transformers} (version 4.21.2) and all models are built using PyTorch \cite{NEURIPS2019_9015} (version 1.12.1).

\textbf{Organization.} As for how the rest of the paper is outlined, \hyperref[sec:data]{\S 2} describes the data used in the training, evaluation, and final testing of our models, \hyperref[sec:methodology]{\S 3} recounts the procedures used in our self-training approach, \hyperref[sec:results]{\S 4} discusses our findings, and \hyperref[sec:conclusion]{\S 5} wraps up the paper with our final remarks and ideas for future direction.

\section{Data}
\label{sec:data}

\subsection{Causal News Corpus}

The CNC dataset \cite{tan-EtAl:2022:LREC} is a corpus of 3,559 event sentences from protest event news labeled on whether a given sentence contains causal relations or not. 
The data of the CNC comes from two workshops focused on mining socio-political data: Automated Extraction of Socio-political Events from News (AESPEN) \cite{hurriyetoglu-etal-2020-automated} in 2020 and the CASE 2021 workshop @ ACL-IJCNLP \cite{hurriyetoglu-etal-2021-challenges}. For the purposes of subtask 1, the data is split into a training set of 2925 examples, a development set of 323 examples, and a final test set of 311 examples that is used as an evaluation benchmark for the competition.

\subsection{Self-labeled Training Data}
\label{sec:wiki}

Sample sentences used in the self-labeling phase of self-training are gathered from 205,328 articles on Wikipedia. The Wikipedia dataset is built from the Wikipedia dump~\footnote{https://dumps.wikimedia.org} and is available as on HuggingFace Dataset library \cite{lhoest-etal-2021-datasets}.
We use the 20220301.simple training split to generate our self-labeled examples.

\section{Methodology}
\label{sec:methodology}

In this section, we review the methods we used in our approach to the sequence causality classification subtask.

\subsection{Self-Training}

We follow a similar teacher-student pipeline as \citealp{yalniz2019billion} that includes using a teacher model to generate a new labeled dataset $\mathcal{D'}$ from the original dataset $\mathcal{D}$ and then training a new student model on both the new labeled dataset $\mathcal{D'}$ and the original dataset $\mathcal{D}$. We use the training split provided of 2925 CNC samples \cite{tan-EtAl:2022:LREC} as the original dataset $\mathcal{D}$, and fine-tune a BERT base-cased model \cite{devlin-etal-2019-bert} for sequence classification, which serves as our teacher model. \hyperref[sec:figure1]{Figure 1a} shows the full pipeline from Wikipedia data collection to saving self-labeled samples.
These self-labeled examples are used as training data for the separate student models later in the experimentation process, as shown in \hyperref[sec:figure1]{Figure 1b}.

% \usepackage{graphicx}
\begin{table*}
\label{sec:table1}
\centering
\resizebox{\textwidth}{!}{%
\begin{tabular}{clclllll}
\hline
\multicolumn{8}{|c|}{\textbf{Baseline Training vs. Self-Training Results}} \\ \hline
\multicolumn{1}{c|}{\multirow{4}{*}{\begin{tabular}[c]{@{}c@{}}Baseline Training\\ (simple fine-tuning, \\ no self-training)\end{tabular}}} &  & \multicolumn{1}{l|}{} & \multicolumn{1}{c}{Accuracy} & \multicolumn{1}{c}{F1} & \multicolumn{1}{c}{Recall} & \multicolumn{1}{c}{Precision} & \multicolumn{1}{c}{MCC} \\ \cline{4-8} 
\multicolumn{1}{c|}{} & BERT & \multicolumn{1}{l|}{} & 0.8204 & 0.8394 & 0.8516 & 0.8276 & 0.6363 \\
\multicolumn{1}{c|}{} & RoBERTa & \multicolumn{1}{l|}{} & 0.8390 & 0.8543 & 0.8561 & 0.8525 & 0.6745 \\
\multicolumn{1}{c|}{} & Google ELECTRA Discriminator & \multicolumn{1}{l|}{} & 0.8365 & 0.8535 & 0.8640 & 0.8432 & 0.6689 \\ \hline
\multicolumn{1}{c|}{\multirow{10}{*}{Self-Training}} & \multicolumn{1}{l|}{} & \multicolumn{1}{c|}{\begin{tabular}[c]{@{}c@{}}Ratio of Positive to Negative Self-Labeled\\ Examples used in training\end{tabular}} & \multicolumn{1}{c}{Accuracy} & \multicolumn{1}{c}{F1} & \multicolumn{1}{c}{Recall} & \multicolumn{1}{c}{Precision} & \multicolumn{1}{c}{MCC} \\ \cline{2-8} 
\multicolumn{1}{c|}{} & \multicolumn{1}{l|}{BERT} & \multicolumn{1}{c|}{1:3} & {\underline{0.8380}} & {\underline{0.8531}} & {\underline{0.8539}} & 0.8525 & 0.6726 \\
\multicolumn{1}{c|}{} & \multicolumn{1}{l|}{} & \multicolumn{1}{c|}{1:1} & 0.8225 & 0.8377 & 0.8315 & 0.8468 & 0.6425 \\
\multicolumn{1}{c|}{} & \multicolumn{1}{l|}{} & \multicolumn{1}{c|}{3:1} & {\underline{0.8380}} & 0.8526 & 0.8502 & {\underline{0.8552}} & {\underline{0.6728}} \\ \cline{2-8} 
\multicolumn{1}{c|}{} & \multicolumn{1}{l|}{RoBERTa} & \multicolumn{1}{c|}{1:3} & 0.8576 & 0.8715 & {\underline{\textbf{0.8764}}} & 0.8671 & 0.7123 \\
\multicolumn{1}{c|}{} & \multicolumn{1}{l|}{} & \multicolumn{1}{c|}{1:1} & {\underline{\textbf{0.8586}}} & 0.8711 & 0.8670 & {\underline{\textbf{0.8755}}} & {\underline{\textbf{0.7149}}} \\
\multicolumn{1}{c|}{} & \multicolumn{1}{l|}{} & \multicolumn{1}{c|}{3:1} & {\underline{\textbf{0.8586}}} & {\underline{\textbf{0.8719}}} & 0.8727 & 0.8711 & 0.7142 \\ \cline{2-8} 
\multicolumn{1}{c|}{} & \multicolumn{1}{l|}{Google ELECTRA Discriminator} & \multicolumn{1}{c|}{1:3} & 0.8400 & 0.8579 & {\underline{\textbf{0.8764}}} & 0.8415 & 0.6760 \\
\multicolumn{1}{c|}{} & \multicolumn{1}{l|}{} & \multicolumn{1}{c|}{1:1} & {\underline{0.8524}} & {\underline{0.8665}} & 0.8689 & {\underline{0.8641}} & {\underline{0.7016}} \\
\multicolumn{1}{c|}{} & \multicolumn{1}{l|}{} & \multicolumn{1}{c|}{3:1} & 0.8421 & 0.8580 & 0.8652 & 0.8510 & 0.6806
\end{tabular}}
\captionsetup{justification=centering}
\caption{Results of the evaluating the CNC development set on both simple fine-tuning with only CNC training data (top) and fine-tuning classifiers on training sets of self-labeled data in addition to CNC training data (bottom). \textbf{Bold} indicates highest performance across all splits and model types, \underline{underline} indicates the highest performance of the specific model type.}
\end{table*}

\subsubsection{Data Preprocessing}

To preprocess Wikipedia data (\S~\ref{sec:wiki}), we first split the articles into individual sentences and discarded all sentences of less than 50 characters and more than 500 characters. 
To self-label the sentences, we feed the sentences into the teacher model and keep all examples with a softmax classifier over a predetermined threshold $\mathcal{T}$. For the purposes of our experiments, we choose a $\mathcal{T}$ of 0.9. In total, we collect a pool of 77,748 positive (causal) examples and 77,940 negative (non-causal) examples.
The large total number of examples collected for this data pool is done to minimize the overlap of examples between the later created self-labeled training splits.

\subsubsection{Training Splits}

From the pools of self-labeled Wikipedia examples, we collect 5 different training sets, all with the size of 10,000 samples but with varying ratios of positive to negative self-labeled examples. We collect sets with positive to negative proportions of 1:3, 1:1, and 3:1 (that is, for a positive to negative proportion of 1:3, we include 2,500 self-labeled positive examples in the training set and 7,500 negative samples).
We design this set-up to test how the different polarity proportions of self-labeled data used in training affect not only overall model accuracy, but also if there is a discrepancy between model precision and recall with the varying polarity splits.
We chose a training split size of 10,000 examples as we notice that self-training performance does not continue to improve with training with splits larger than this \footnote{Observed in our initial internal testing phase}. 
When formulating each set, we randomly reshuffle the positive and negative self-labeled sets and chose the first \textit{s} and \textit{t} positive and negative samples for a training set that require \textit{s} positive examples and \textit{t} negative examples. From there, we combine the \textit{s} positives and the \textit{t} negatives and again shuffle the concatenated training set.

\subsubsection{Fine-tuning on Self-labeled data}

For each self-labeled dataset, we fine-tune a classifier---which serves as our student model---on one epoch of the self-labeled dataset and then five epochs of the CNC provided training data. The predictions generated after the final epoch of training are used for evaluation. We run our experiments with student classifiers built on BERT base-cased \cite{devlin-etal-2019-bert}, RoBERTa base \cite{liu2019roberta}, and Google ELECTRA-base-discriminator \cite{clark2020electra} pre-trained models.

\subsection{Transformer-based Data Augmentation and Multi-task Learning}

In our participation of the CASE workshop, we also explore both Transformer-based data augmentation and multi-task learning as a means to improve performance on causality classification. 
While our both of these approaches are out-performed by our self-training approaches and thus are not the main focus of this paper, we still find significant results with these methods and implement both a Transformer-based data augmentation technique and a multi-task architecture that comprehensively outperform the baseline classifier for the given task.
The full methodology and experimentation of our Transformer-based data augmentation and multi-task learning approaches are available in the \hyperref[sec:appendix]{Appendix}.

\section{Experiments and Results}
\label{sec:results}

\subsection{Experiment Set up}

In our experimentation setup, we test all three backbone models (BERT, RoBERTa, and Google ELECTRA Discriminator) with both the self-training pipeline and a simple fine-tuning process that only uses the provided CNC training set that served as the baseline. In the baseline experiments, the classifiers are trained solely on five epochs of the CNC training data.
We conduct five trials of each setup, each trial having a randomly initialized seed.
We use the CNC development set as our testing benchmark due to the limited number of allowed workshop testing phase submissions.

\subsection{Classifier Set up}

In our experiments, we run all trials on a Tesla V100-SXM2-16GB GPU device. We use an AdamW optimizer with $\beta_1$ = 0.9, $\beta_2$ = 0.999, a learning rate of $5e-5$, and a linear decay rate. Finally, all experiments are run with a batch size of 8.

\subsection{Findings}

\hyperref[sec:table1]{Table 1} displays the results from our experiments, which include the averages of 5 trials for each set-up. 
From the table, we can see that every self-training setup outperforms the baseline classifier in terms of accuracy, with an average accuracy improvement of 1.33\% across all models and polarity splits. Furthermore, for all but one self-training set-up, there is an improvement of the F$_1$ score from the baseline, with an average improvement of 0.011.

Other key takeaways from our results are that
1) there is very little overall performance degradation across the polarity splits (1:3, 1:1, 3:1) in the self-labeled datasets (only the BERT model shows a range of F1 scores above 0.01) and 2) there is low discrepancy between recall and precision among the splits (only the 1:3 split with an ELECTRA backbone shows a recall-precision discrepancy > 0.015.)

\subsection{Competition Results}

Our best-performing prediction set of the final competition testing comes from a RoBERTa classifier trained on a self-labeled training set with a polarity ratio of 1:1. 
The results of our all of our competition submissions \footnote{Workshop competition limited participants to five submissions for the testing phase} are shown in \hyperref[sec:submission]{Table 2}. 
All of our competition submissions comprehensively outperform the provided baseline, and our best overall performing submission achieve competition rankings of 6$^{th}$ in accuracy, 10$^{th}$ in F1, 7$^{th}$ in recall, 7$^{th}$ in precision, and 10$^{th}$ in MCC.

% \usepackage{graphicx}
\begin{table*}
\label{sec:submission}
\centering
\resizebox{\textwidth}{!}{%
\begin{tabular}{lclllll}
\hline
\multicolumn{7}{|c|}{\textbf{Competition Results (CNC Test Set)}} \\ \hline
\multicolumn{1}{l|}{} & \multicolumn{1}{c|}{\begin{tabular}[c]{@{}c@{}}Ratio of Positive to Negative \\ Self-Labeled Examples\end{tabular}} & \multicolumn{1}{c}{Accuracy} & \multicolumn{1}{c}{F1} & \multicolumn{1}{c}{Recall} & \multicolumn{1}{c}{Precision} & \multicolumn{1}{c}{MCC} \\ \hline
\multicolumn{1}{l|}{RoBERTa} & \multicolumn{1}{c|}{1:3} & 0.8071 & 0.8256 & 0.8068 & 0.8452 & 0.6108 \\
\multicolumn{1}{l|}{} & \multicolumn{1}{c|}{1:1} & \textbf{0.8135} & \textbf{0.8398} & \textbf{0.8636} & 0.8172 & 0.6185 \\
\multicolumn{1}{l|}{} & \multicolumn{1}{c|}{3:1} & 0.7974 & 0.8215 & 0.8239 & 0.8192 & 0.5873 \\ \hline
\multicolumn{1}{l|}{\multirow{2}{*}{ELECTRA}} & \multicolumn{1}{c|}{1:1} & \textbf{0.8135} & 0.8324 & 0.8181 & \textbf{0.8471} & \textbf{0.6228} \\
\multicolumn{1}{l|}{} & \multicolumn{1}{c|}{3:1} & 0.7942 & 0.8107 & 0.7784 & 0.8457 & 0.5886 \\ \hline
\multicolumn{2}{l|}{\begin{tabular}[c]{@{}l@{}}Provided Competition Baseline\\ (BERT baseline model)\end{tabular}} & 0.7781 & 0.8120 & 0.8466 & 0.7801 & 0.5452
\end{tabular}
}
\captionsetup{justification=centering,margin=1cm}
\caption{Results of competition submissions on CNC test set. \textbf{Bold} indicates highest performer.}
\end{table*}

\section{Conclusion and Discussion}
\label{sec:conclusion}

This paper explores how training a classifier on self-labeled data can improve the performance of sequence classification tasks. In our case, we examine the effect of self-training on the task of event causality in socio-political event data as part of Subtask 1 of Shared Task 3 of the CASE workshop @ EMNLP 2022.

Our results show that training a classifier on self-labeled data using a teacher-student approach comprehensively improves task performance. Furthermore, we find that performance improvement from self-training did not differ significantly between self-labeled training sets with varying levels of example polarity. 
This indicates that the model is capable of reaping the full benefits of self-training despite having limited access to positive or negative samples. One thing that could help explain this is our relatively high threshold $\mathcal{T}$ of 0.9 which determines whether or not to keep an example during the initial self-labeled process. Further research should explore whether a lower $\mathcal{T}$ could alter the benefits of self-training, especially when self-labeled examples would have a higher chance of being incorrectly labeled.

Next, given that our self-labeled examples are gathered from an assortment of articles from Wikipedia, it should be well noted that the benefits of self-training are apparent even when the self-labeled examples are not domain specific to the original labeled data.
We decide to use Wikipedia as the source of our self-labeled examples as we view it as a more accessible source with far greater amounts available unlabeled data.
Thus, our findings indicate that performance improvements from self-training work with non-domain specific data, which alleviates us from the restriction of confining our self-labeled data to the single domain of the original labeled data.

Finally, one more aspect of our experiments that should be further explored is the classifier's actual dependence on the self-labeled data versus the originally provided training data. 
In our setup, we choose to train our models on one epoch of self-labeled data and then on five epochs of the original training data in order to prioritize the true labeled training data. We believe that it would be worthwhile to explore training classifiers with a higher training priority on the self-labeled data, or even to test the performance of classifiers trained solely on the self-labeled data, without the original true data.

\section*{Acknowledgments}

The work for this study is supported by the National Key R\&D Program of China (2020AAA0105200) and the Sam '75 and Meg Woodside Fund for Career Exploration provided through Carleton College.

% % Entries for the entire Anthology, followed by custom entries
\bibliography{custom}
\bibliographystyle{acl_natbib}

% \newpage 

\section{Appendix}
\label{sec:appendix}

\begin{figure*}
    \label{sec:figure2}
    \centering
    \includegraphics[width=15cm]{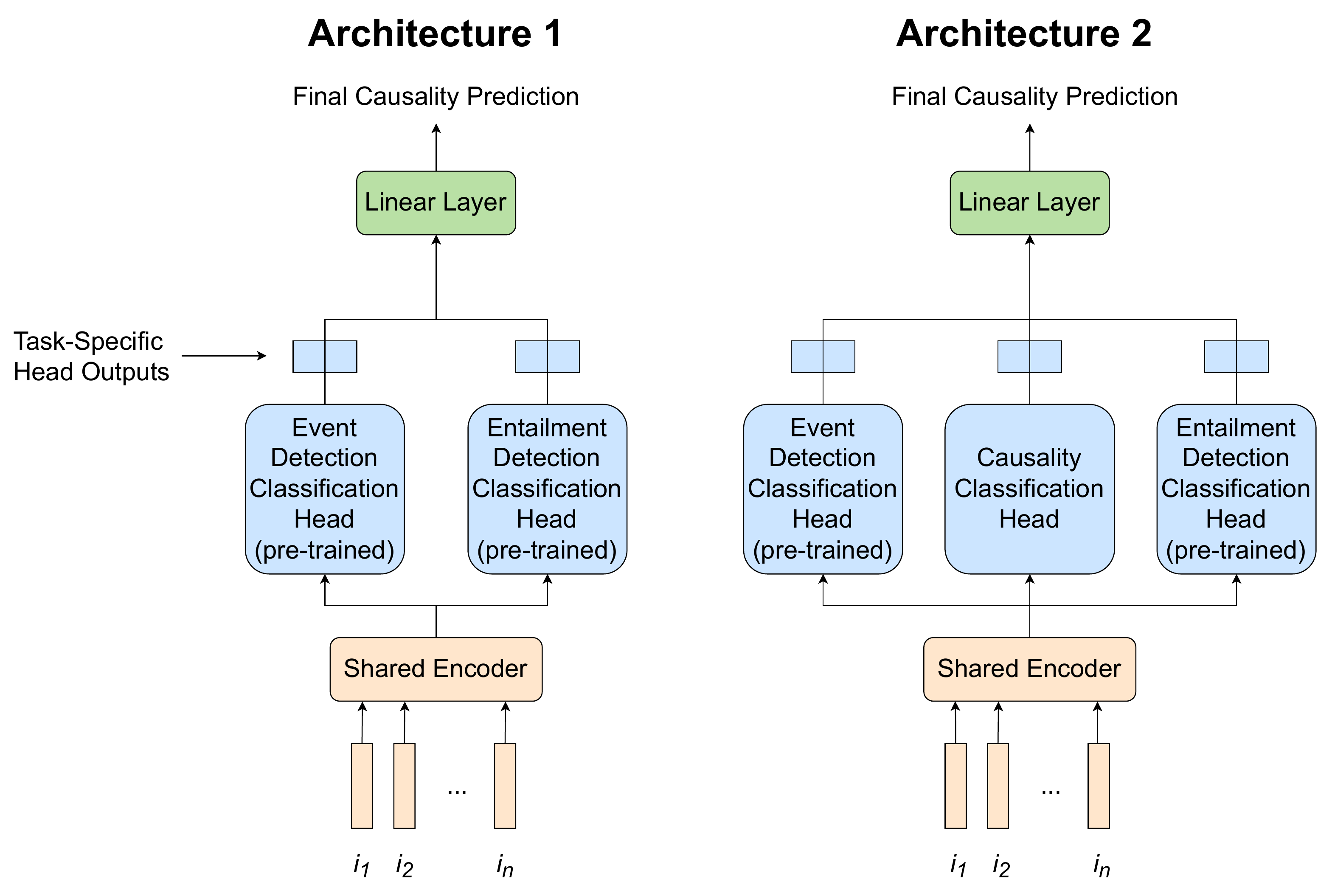}
    \caption{Multi-task learning architectures used in supplementary testing.}
\end{figure*}

Here, we outline the supplementary experimentation we conducted to compare our self-training results with other methods we explored in the CASE event causality competition. These methods include a few popular transformer-based textual data augmentation techniques and two multi-task learning-based classifier architectures.

\subsection{Transformer-based Data Augmentation}

In general, data augmentation---within the context of textual data---works by altering a given labeled example and attaching the label of the original example to the augmented one. 
Each of the transformer-based data augmentation techniques is considered with the same goal of increasing the training data space to improve the model performance on the task of causality classification. We use the CNC training split of 2925 as the original data to be augmented in our experiments.

\subsubsection{Sequence to Sequence Data Augmentation}

Sequence-to-sequence text augmentation works by taking the sentence of the original example (all of our data examples are English examples), translating the sentence into a foreign language, and then finally translating the rendered sentence back to the original language. This works by altering some words or clusters of words in a sentence while preserving the original structure and semantics. For the purposes of our experiments, we use two foreign languages to augment the data, German and Russian, using HuggingFace's ported versions of the Facebook FAIR’s WMT19 News Translation Task Submission \cite{ng2019facebook}. 
The sequence-to-sequence augmented training set has 8,775 examples; 2,925 from the original training set and 5,850 augmented examples.

\subsubsection{Random Fill-mask Data Augmentation}

In random fill-mask augmentation, we first randomly select a word from the original. From there, we replace the selected word with a masking token and use the new sentence with masking as input to a pre-trained RoBERTa fill-mask language model \cite{liu2019roberta} to select the three most likely fill-mask options for the masked word. 
With the three selected substitutions for the masked word, we create three new sentences by replacing each respective substitution with the original masked word and keeping the original label of the sentence with the new augmented examples. The final random fill-mask augmented set has 11,700 total samples.

\subsubsection{NER Fill-Mask Data Augmentation}

The NER fill-mask data augmentation functions in a similar fashion to the random fill-mask data augmentation, but instead of selecting a single random word to replace, we make substitutions to any named entities identified by Named Entity Recognition (NER) \cite{mikheev1999named, mohit2014named}. Specifically, we use the EntityRecognizer module from spaCy~\footnote{https://spacy.io/} to identify which tokens in a sentence corresponded to named entities. For each example sentence from the original training data that contained named entities, we create three augmented sentences by substituting the best unused fill-mask option for each named entity in the text. The final NER augmented dataset has 10,443 example sentences in total.

\subsection{Multi-Task Learning Approaches}

\begin{table*}
\label{sec:table3}
\centering
\begin{tabular}{llrrrrr}
\hline
\multicolumn{7}{|c|}{\textbf{Supplementary Experiments Results}} \\ \hline
 &
  \multicolumn{1}{c|}{} &
  \multicolumn{1}{c}{Accuracy} &
  \multicolumn{1}{c}{F1} &
  \multicolumn{1}{c}{Recall} &
  \multicolumn{1}{c}{Precision} &
  \multicolumn{1}{c}{MCC} \\ \hline
\multicolumn{2}{l|}{Baseline} &
  0.8390 &
  0.8543 &
  0.8561 &
  0.8525 &
  0.6745 \\
\multicolumn{2}{l|}{Self-Training (1:1 polarity)} &
  {\color[HTML]{212121} \textbf{0.8586}} &
  {\color[HTML]{212121} \textbf{0.8711}} &
  {\color[HTML]{212121} \textbf{0.8670}} &
  {\color[HTML]{212121} \textbf{0.8755}} &
  {\color[HTML]{212121} \textbf{0.7149}} \\ \hline
\multicolumn{1}{l|}{} &
  \multicolumn{1}{l|}{Sequence to Sequence} &
  0.8235 &
  0.8430 &
  \textbf{0.8596} &
  0.8270 &
  0.6424 \\
\multicolumn{1}{l|}{} &
  \multicolumn{1}{l|}{Random Fill-Mask} &
  \textbf{0.8406} &
  \textbf{0.8562} &
  \textbf{0.8574} &
  \textbf{0.8556} &
  \textbf{0.6778} \\
\multicolumn{1}{l|}{\multirow{-3}{*}{\textbf{\begin{tabular}[c]{@{}l@{}}Transformer-based \\ Data Augmentations\end{tabular}}}} &
  \multicolumn{1}{l|}{NER Fill-Mask} &
  \textbf{0.8452} &
  \textbf{0.8571} &
  0.8427 &
  \textbf{0.8721} &
  \textbf{0.6888} \\ \hline
\multicolumn{1}{l|}{} &
  \multicolumn{1}{l|}{Architecture 1} &
  \textbf{0.8498} &
  \textbf{0.8655} &
  \textbf{0.8764} &
  \textbf{0.8548} &
  \textbf{0.6960} \\
\multicolumn{1}{l|}{\multirow{-2}{*}{\textbf{Multi-Task Learning}}} &
  \multicolumn{1}{l|}{Architecture 2} &
  0.8313 &
  0.8489 &
  \textbf{0.8596} &
  0.8385 &
  0.6583
\end{tabular}
\captionsetup{justification=centering}
\caption{Results from supplementary testing done on CNC development set. All runs use a RoBERTa backbone model. The baseline and self-training results are taken from the main experiments of the paper. \textbf{Bold} indicates outperforming the baseline.}
\label{table1}
\end{table*}

Multi-task learning (MTL) \cite{caruana1997multitask,zhang2021survey,ruder2017overview} is a paradigm of machine learning that improves the performance of a model in a given task by leveraging simultaneous learning of other distinct but related tasks. Our MTL architectures learn the distinct tasks of entailment classification (binary classification of whether the meaning of one sentence can be inferred from another sentence) and event detection (whether a sentence contains information about a socio-political event), then combine the prior knowledge of those two tasks to help supplement the classifier's prediction to the task of causality classification.

\subsubsection{MTL Datasets}

We used two distinct datasets for the multi-task learning of entailment detection and event detection.

\textbf{Entailment Detection Dataset}
\label{sec:entailment}
We evaluate using the Recognizing Textual Entailment (RTE) task provided in the GLUE Benchmark~\cite{wang2018glue} for the entailment detection task. 
In training, we used the given training set that consisted of 2490 examples. Each example from the RTE dataset consisted of two sentences and a binary label on whether or not one of the two sentences holds logical entailment to the other. To better fit the structure of the other data, we concatenated the two provided sentences into a single text to be used as input into the models.

\textbf{Event Detection Dataset}
\label{sec:event detection}
In order to learn the task of event detection, we used data provided in the second shared task of CASE @ ACL-IJCNLP 2021~\cite{hurriyetoglu-etal-2021-challenges}, which provided data to the object of sentence-level event classification. 
The data provided from subtask 2 of CASE 2021 included 1023 examples sentences of socio-political events, labeled using the Armed Conflict Location \& Event Data Project (ACLED)~\cite{raleigh2010introducing} event taxonomy, which consists of 25 fine-grained event subtypes. 
These 1023 example sentences are concatenated with 720 non-event-specific English sentences to create an event detection dataset, with all sentences coming from the event classification receiving a label of '1', denoting that the sentence contained information about an event.

\subsubsection{MTL Pre-training}

Prior to fine-tuning our models for the task of causality classification, we train a shared encoder \cite{guo2021safe}---a RoBERTa pre-trained model---on the separate tasks of event detection and entailment detection by fine-tuning the shared encoder on the respective datasets for each task. We fine-tune three epochs for both tasks.

\subsubsection{MTL Architectures}

We experiment with two similar but different architectures in MTL testing. In both architectures, we first simultaneously fine-tune a classifier on the two tasks of entailment detection and event detection. Because we have distinct datasets for each respective task, we implement this by using the shared encoder approach, where model parameters are hard-shared and each task has its own task-specific classification head.

The distinction between our two MTL architectures comes from how we choose to combine prior knowledge. The architectures we build are shown in \hyperref[sec:figure2]{Figure 2}. Both architectures include task-specific classification heads for the tasks of entailment detection and event detection. 
The distinction between the two architectures comes in where Architecture no. 2 also includes a causality-specific classification head; the outputs of all three task heads are combined and inputted into a final linear layer to output the final logits prediction. Architecture no. 1 omits the causality-specific classification head and simply combines the outputs of the entailment detection and event detection heads before the linear layer.

\subsection{Supplementary Experiments and Results}

\subsubsection{Set up}

For the supplementary experiments, we follow the same setup as in the main study to maintain consistency. Thus, models trained on a transformer-augmented dataset are trained on five epochs of the respective dataset, and each MTL architecture is trained on five epochs of the CNC training set. The evaluations are calculated on the predictions made after the final epoch of training. Likewise, we use the same hyperparameter setup as the main experiments, meaning that we run all trials on a Tesla V100-SXM2-16GB GPU device. Hyperparameters are listed in \hyperref[sec:hypers]{Table 4}. For purposes of the supplementary experiments, we run all trials using a RoBERTa backbone.

\begin{table}
\label{sec:hypers}
\centering
\begin{tabular}{lr}
\multicolumn{2}{c}{AdamW Optimizer w/ Linear Decay} \\ \hline
\multicolumn{1}{l|}{$\beta_1$}              & 0.9   \\
\multicolumn{1}{l|}{$\beta_2$}              & 0.999 \\
\multicolumn{1}{l|}{Per device batch size}  & 8    
\end{tabular}
\caption{Classifier hyperparameter settings.}
\label{table2}
\end{table}

\subsubsection{Results}

\hyperref[sec:table3]{Table 3} displays the results of our supplementary tests. Consistent with the main study, the results are the averages over five trials for each of the setups on the CNC development set. Between the transformer-augmented experiments, the random fill-mask and NER fill-mask experiments outperformed the baseline in terms of both accuracy and F$_1$ score. Similarly, Architecture no. 1 of the MTL approaches also outperformed the baseline in terms of accuracy and F$_1$.

\subsubsection{Discussion}

We include the supplementary experiments to \textbf{1)} show how our self-training results compared to popular state-of-the-art data augmentation techniques using contemporary NLP, and \textbf{2)} propose the multi-task learning architectures we originally developed for the Subtask 1 of the competition. Although the final results of the MTL approaches did not reach the same level of performance as the self-training approaches and therefore did not belong in the main paper, we believe the MTL experiments and results are still notable and worth mentioning for further investigation.

\end{document}